\documentclass[twoside]{article}

\usepackage[accepted]{aistats2023}

\usepackage[round]{natbib}

\let\originalleft\left
\let\originalright\right
\renewcommand{\left}{\mathopen{}\mathclose\bgroup\originalleft}
\renewcommand{\right}{\aftergroup\egroup\originalright}

\usepackage[acronyms, shortcuts]{glossaries}
\glsdisablehyper
\usepackage{subcaption}
\usepackage{paralist}
\usepackage{multibib}
\newcites{app}{Appendix References}
\usepackage{hyperref}

\usepackage{xspace}
\usepackage{multicol}

\usepackage{capt-of}
\usepackage{graphicx}
\usepackage{booktabs} %

\usepackage{amsmath}
\usepackage{amssymb}
\usepackage{mathtools}
\usepackage{amsthm}

\usepackage{siunitx}
\renewcommand{\cite}[1]{\citep{#1}}

\usepackage{amssymb} %
\usepackage{amsmath}
\usepackage{bm} %

\newcommand{\eq}[2][my_equation]{\begin{equation}\label{eq:#1}#2\end{equation}}
\newcommand{\al}[2][my_equation]{\begin{align}\label{eq:#1}#2\end{align}}
\newcommand{\als}[2][my_equation]{\begin{align}\label{eq:#1}\begin{split}#2\end{split}\end{align}}

\providecommand\refeq{} %
\renewcommand{\refeq}[1]{Eq.~\ref{eq:#1}}

\newcommand{\reffig}[1]{Figure~\ref{fig:#1}}
\newcommand{\reftab}[1]{Table~\ref{tab:#1}}
\newcommand{\refsec}[1]{section~\ref{sec:#1}}

\newcommand{\loss}{\mathcal{L}}

\newcommand{\set}[1]{\left\{ #1 \right\}}
\newcommand{\setsize}[1]{\left| #1 \right|}
\newcommand{\R}{\mathbb{R}} %

\renewcommand{\b}[1]{\bm{#1}} %

\newcommand{\indicator}[1]{\mathbb{I}_{\left\{ #1 \right\}}}

\DeclareMathOperator{\diag}{diag}

\newcommand{\transpose}{\intercal}

\newcommand{\uniform}[1]{\mathcal{U}\left(#1\right)}

\newcommand{\papertitle}{Temporal Graph Neural Networks for Irregular Data}

\newcommand{\citeauth}[1]{\citeauthor{#1} (\citeyear{#1})}
\newcommand{\refapp}[1]{appendix~\ref{sec:#1}}

\newcommand{\G}{\mathcal{G}}
\newcommand{\obsn}{\mathcal{O}}

\newcommand{\neigh}[1]{\mathcal{N}(#1)}
\newcommand{\neighset}[1]{\set{#1}_{m \in \neigh{n}}}
\newcommand{\edgew}[2]{e_{#1,#2}}
\newcommand{\Nobs}{N_{\text{obs}}}

\DeclareMathOperator{\gnn}{GNN}

\newcommand{\dy}{d_y}
\newcommand{\dx}{d_x}
\renewcommand{\dh}{d_h}
\newcommand{\pred}[2][n]{\b{\hat{y}}^{#1}_{#2}}

\newcommand{\latstate}[2][n]{\b{h}^{#1}(#2)}
\newcommand{\latent}[1][n]{\b{h}^{#1}}
\newcommand{\gruout}[2][n]{\b{\hat{h}}^{#1}_{#2}}
\newcommand{\target}[2][n]{\b{\bar{h}}^{#1}_{#2}}
\newcommand{\hode}[2][n]{\b{\tilde{h}}^{#1}(#2)}
\newcommand{\decweight}[2][n]{\b{\omega}^{#1}_{#2}}
\newcommand{\obsvec}{\b{y}_i^n}
\newcommand{\featurevec}[1][i]{\b{x}_{#1}^n}

\newcommand{\gruparth}[1]{\b{u}_{i,#1}^n}
\newcommand{\grupartx}[1]{\b{v}_{i,#1}^n}

\newcommand{\gruline}[1]{\sigma(\grupartx{#1} + \gruparth{#1} + \b{b}_{#1})}
\newcommand{\gruinput}[1][n]{\b{\tilde{x}}_i^{#1}}
\newcommand{\grur}{\b{r}_i^n}
\newcommand{\gruz}{\b{z}_i^n}
\newcommand{\gruh}{\b{q}_i^n}
\newcommand{\grurt}{\b{\bar{r}}_i^n}
\newcommand{\gruzt}{\b{\bar{z}}_i^n}
\newcommand{\gruht}{\b{\bar{q}}_i^n}

\newcommand{\futureloss}[1]{\loss_{#1}}
\newcommand{\lmse}{\futureloss{\text{MSE}}}
\newcommand{\yar}{\hat{\b{y}}_{i \rightarrow j}^m}
\newcommand{\ninit}{N_\text{init}}
\newcommand{\nmax}{N_\text{max}}

\DeclareMathOperator{\lastobsi}{prev}

\newcommand{\grubase}{\hbox{GRU-D}\xspace}
\newcommand{\grujoint}{\grubase (joint)\xspace}
\newcommand{\grunode}{\grubase (node)\xspace}
\newcommand{\transbase}{Transformer\xspace}
\newcommand{\transjoint}{\transbase (joint)\xspace}
\newcommand{\transnode}{\transbase (node)\xspace}
\newcommand{\itgnn}{TGNN4I\xspace}

\newcommand{\predprev}{Predict Previous\xspace}
\newcommand{\lgode}{LG-ODE\xspace}

\newcommand{\bay}{PEMS-BAY\xspace}
\newcommand{\la}{METR-LA\xspace}

\newcommand{\tmin}{T_{\min}}
\newcommand{\tmax}{T_{\max}}

\newcommand{\simbase}[1][n]{\rho^{#1}(t)}
\newcommand{\simsignal}[2][n]{\kappa^{#1}(#2)}
\newcommand{\simfull}[1][n]{y^{#1}(t)}
\newcommand{\simnoise}[1][n]{\epsilon^{#1}(t)}

\newcommand{\sinangle}[1]{\theta_{#1}}
\newacronym{RNN}{RNN}{Recurrent Neural Network}
\newacronym{GRU}{GRU}{Gated Recurrent Unit}
\newacronym{GNN}{GNN}{Graph Neural Network}
\newacronym{RMSE}{RMSE}{Root Mean Squared Error}
\newacronym{MSE}{MSE}{Mean Squared Error}
\newacronym{ODE}{ODE}{Ordinary Differential Equation}
\newacronym{DAG}{DAG}{Directed Acyclic Graph}
\newacronym{ELBO}{ELBO}{Evidence Lower Bound}
\newacronym{TGNN4I}{TGNN4I}{Temporal Graph Neural Network for Irregular data}
\newacronym{USHCN}{USHCN}{United State Historical Climatology Network}

\usepackage{multirow}

\begin{document}

\twocolumn[
\aistatstitle{\papertitle}
\aistatsauthor{
Joel Oskarsson
\And
Per Sid\'en
\And
Fredrik Lindsten
}
\aistatsaddress{
Link\"{o}ping University
\And  
Link\"{o}ping University\\
Arriver Software AB
\And 
Link\"{o}ping University
} ]

\begin{abstract}
This paper proposes a temporal graph neural network model for forecasting of graph-structured irregularly observed time series.
Our TGNN4I model is designed to handle both irregular time steps and partial observations of the graph.
This is achieved by introducing a time-continuous latent state in each node, following a linear Ordinary Differential Equation (ODE) defined by the output of a Gated Recurrent Unit (GRU).
The ODE has an explicit solution as a combination of exponential decay and periodic dynamics.
Observations in the graph neighborhood are taken into account by integrating graph neural network layers in both the GRU state update and predictive model.
The time-continuous dynamics additionally enable the model to make predictions at arbitrary time steps.
We propose a loss function that leverages this and allows for training the model for forecasting over different time horizons.
Experiments on simulated data and real-world data from traffic and climate modeling validate the usefulness of both the graph structure and time-continuous dynamics in settings with irregular observations.

\end{abstract}

\section{INTRODUCTION}
Many real-world systems can be modeled as graphs.
When data about such systems is collected over time, the resulting time series has additional structure induced by the graph topology.
Examples of such temporal graph data is the traffic speed in the road network \cite{dcrnn_traffic} and the spread of disease in different regions \cite{chickenpox}.
Building accurate machine learning models in this setting requires taking both the temporal and graph structure into account.

While many works have studied the problem of modeling temporal graph data \cite{gnn_survey}, these approaches generally assume a constant sampling rate and no missing observations.
In real data it is not uncommon to have irregular or missing observations due to non-synchronous measurements or errors in the data collection process.
Dealing with such irregularities is especially challenging in the graph setting, as node observations are heavily interdependent.
While observations in one node could be modeled using existing approaches for irregular time series \cite{latent_node, cru}, the situation becomes complicated when irregular observations in different nodes occur at different times.

In this paper we tackle two kinds of irregular observations in graph-structured time series:
\begin{inparaenum}[(1)]
    \item irregularly spaced observation times, and
    \item only a subset of nodes being observed at each time point.
\end{inparaenum}
We propose the \itgnn model for time series forecasting.
The model uses a time-continuous latent state in each node, which allows for predictions to be made at any time point.
The latent dynamics are motivated by a linear \gls{ODE} formulation, which has a closed form solution.
This \gls{ODE} solution corresponds to an exponential decay \cite{decay_gru} together with an optional periodic component.
New observations are incorporated into the state by a \gls{GRU} \cite{gru}.
Interactions between the nodes are captured by integrating \gls{GNN} layers both in the latent state updates and predictive model.
To train our model we introduce a loss function that takes into account the time-continuous model formulation and irregularity in the data.
We evaluate the model on forecasting problems using traffic and climate data of varying degrees of irregularity and a simulated dataset of periodic signals.

\section{RELATED WORK}
\glspl{GNN} are deep learning models for graph-structured data \cite{mpnn, gnn_survey}.
By learning representations of nodes, edges or entire graphs \glspl{GNN} can be used for many different machine learning tasks.
These models have been successfully applied to diverse areas such as 
weather forecasting \cite{graphcast}
molecule generation \cite{moflow} and
video classification \cite{graphvid}.
Temporal \glspl{GNN} model also time-varying signals in the graph.
This extension to graph-structured time series is achieved by combining \gls{GNN} layers with recurrent \cite{dcrnn_traffic}, convolutional \cite{graph_wavenet, st_gcn} or attention \cite{attention_stgcn} architectures.
While the graph is commonly assumed to be known a priori, some approaches also explore learning the graph structure jointly with the temporal \gls{GNN} model \cite{raindrop, connecting_dots}.

Time series forecasting is a well-studied problem and a vast amount of methods exist in the literature.
Traditional methods in the area include ARIMA models, vector auto-regression and Gaussian Processes \cite{timeseries, gps_for_timeseries}.
Many deep learning approaches have also been applied to time series forecasting.
This includes \glspl{RNN} \cite{forecasting_with_python}, temporal convolutional neural networks \cite{forecasting_tcnn} and Transformers \cite{transformers_forecasting}.

The latent state of an \gls{RNN} can be extended to continuous time by letting the state decay exponentially in between observations \cite{decay_gru}.
Such decay mechanisms have been used for modeling data with missing observations \cite{decay_gru}, doing imputation \cite{brits} and parametrizing point processes \cite{neural_hawkes}.
Another way to define time-continuous states is by learning a more general \gls{ODE}.
In neural \gls{ODE} models \cite{neural_ode, on_neural_odes} the latent state is the solution to an \gls{ODE} defined by a neural network.
Neural \glspl{ODE} have been successfully applied to irregular time series \cite{latent_node} and can be used to define the dynamics of temporal \glspl{GNN} \cite{st_graph_ode, cont_depth_graph_ode}.
\citeauth{cont_depth_graph_ode} use such a model for graph-structured time series with irregular time steps, but consider the full graph to be observed at each observation time.
Also the GraphCON framework of \citeauth{oscillator_networks} combines \glspl{GNN} with a second order system of \glspl{ODE}.
In \hbox{GraphCON} the time-axis of the \gls{ODE} is however aligned with the layers of the \gls{GNN}.
This makes the framework suitable for node- and graph-level predictive tasks, rather than time-series modeling.

Another related body of work is concerned with using \glspl{GNN} for data imputation in time series \cite{filling_the_gaps, tsi_gnn, football_imputation}.
These methods generally do not assume that the time series come with some known graph structure.
Instead, the \gls{GNN} is defined on some graph specifically constructed for the purpose of performing imputation.

The closest work to ours found in the literature is the \lgode model of \citeauth{lgode}.
They consider the same types of irregularities, but are motivated more by a multi-agent systems perspective.
\lgode is based on an encoder-decoder architecture and trained by maximizing the \gls{ELBO}.
The encoder builds a spatio-temporal graph of observations and aggregates information using an attention mechanism \cite{attention_all_you_need}.
The decoder then extrapolates to future times by solving a neural ODE.
The encoder-decoder setup differs from our \itgnn model that sequentially incorporates observations and auto-regressively makes predictions at every time point.
Because of the multi-agent motivation \citeauth{lgode} are also more focused on smaller graphs with few interacting entities, but longer forecasting horizons.
An extended version of \lgode, called CG-ODE \cite{cgode}, also aims to learn the graph structure in the form of a weighted adjacency matrix.

\section{A TEMPORAL GNN FOR IRREGULAR OBSERVATIONS}
\label{sec:method}
\begin{figure*}[t]
\begin{center}
\includegraphics[width=\linewidth]{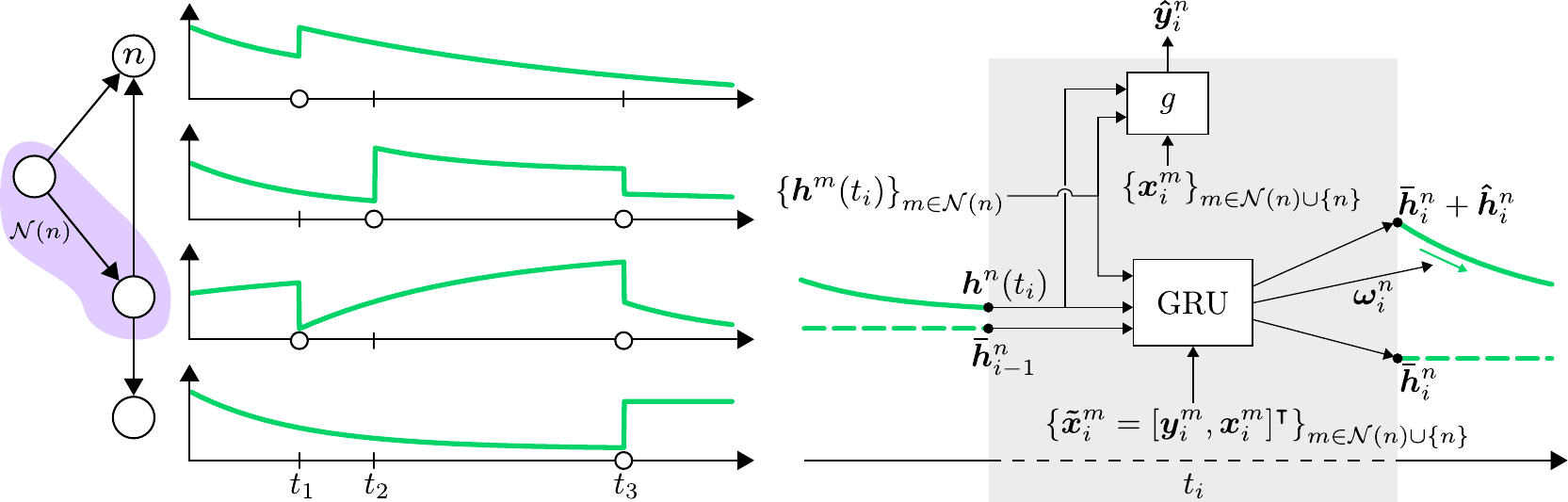}
\caption{
\textbf{Left:}
Example graph with four nodes and their latent states, here one-dimensional for illustration purposes. 
Node observations are indicated with $\bigcirc$.
\textbf{Right:}
Schematic diagram of the GRU update and predictive model $g$. %
Everything inside the shaded area happens instantaneously at time $t_i$.
The \gls{GRU} cell outputs the new initial state $\target{i} + \gruout{i}$, the static component $\target{i}$ and the \gls{ODE} parameters $\decweight{i}$.
The parameters $\decweight{i}$ define the dynamics until the next observation.
The prediction $\pred{i}$ here is based on information from all earlier time points.
}
\label{fig:model_illustration}
\end{center}
\end{figure*}

\subsection{Setting}
\label{sec:setting}
Consider a directed or undirected graph $\G = (V, E)$ with node set $V$ and edge set $E$.
Let $\set{t_i}_{i=1}^{N_t}$ be a set of (possibly irregular) time points s.t. $0 < t_1 < \dots < t_{N_t}$.
We will here present our model for a single time series, but in general we have a dataset containing multiple time series.
Let $\obsn_{i} \subseteq V$ be the set of nodes observed at time $t_i$. 
If $n \in \obsn_{i}$, we denote the observed value as $\obsvec \in \R^{\dy}$ and any accompanying input features as $\featurevec \in \R^{\dx}$.
We let $\obsvec = \featurevec = \b{0}$ if $n \notin \obsn_{i}$.
Note that this general setting encompasses a spectrum of irregularity, from single node observations ($\setsize{\obsn_i} = 1~\forall i$) to fully observed graphs ($\obsn_i = V~\forall i$).
A table of notation is given in \refapp{notation}.%

The problem we consider is that of forecasting.
At future time points we want to predict the value at each node, given all earlier observations.
Since observations are irregular and we want to make predictions at arbitrary times, we need to consider models that can make predictions for any time in the future.

We consider a model where at each node $n$ a latent state $\latstate{t} \in \R^{\dh}$ evolves over continuous time.
We define the dynamics of $\latstate{t}$ by:
\begin{inparaenum}[(1)]
    \item how $\latstate{t}$ evolves in between observations, and
    \item how $\latstate{t}$ is updated when node $n$ is observed.
\end{inparaenum}
If node $n$ is observed at time $t_i$ we incorporate this observation into the latent state using a GRU cell \cite{gru}.
This information can then be used for making predictions at future time points.
An overview of our model is given in \reffig{model_illustration}.

\subsection{Time-continuous Latent Dynamics}
\label{sec:dynamics}
Consider a time interval $]t_i, t_j]$ where node $n$ is not observed.
During this interval we define the latent state of node $n$ by the sum $\latstate{t} = \target{i} + \hode{t}$.
The first part $\target{i}$ is  constant over the time interval, constituting a base level around which the state evolves.
The dynamics of $\hode{t}$ are dictated by a linear \gls{ODE} of the form
\eq[linear_ode]{
    d\hode{t} = A\hode{t} ~ dt
}
with $A \in \R^{\dh \times \dh}$ and initial condition $\hode{t_i} = \gruout{i}$.
Over this interval the \gls{ODE} has a closed form solution \cite{dynamical_systems} given by
\eq[ode_solution]{
    \hode{t} = \exp\left( \delta_t A\right) \gruout{i}
}
where $\exp$ is the matrix exponential function and we define $\delta_t = t - t_i$.
Assuming that all eigenvalues of $A$ are unique, we can use its eigen-decomposition\footnote{Recall that the eigen-decomposition of a diagonalizable matrix $A$ is given by $A=Q \Lambda Q^{-1}$, where $\Lambda$ is a diagonal matrix containing the eigenvalues of $A$ and the columns of $Q$ are the corresponding eigenvectors \cite{matrix_algebra}.}  
$A = Q \Lambda Q^{-1}$ to write
\eq[ode_solution_eigen]{
    \hode{t} = Q\exp\left(\delta_t \Lambda \right)Q^{-1} \gruout{i}.
}
Since $Q$ contains an eigen-basis of $\R^{\dh}$ it can be viewed as a transition matrix, changing the basis of the latent space.
While we could in principle learn $Q$, we note that the basis of the latent space has no physical interpretation and we can without loss of generality choose it such that $Q=I$.

Next, if we make the assumption that $A$ has real eigenvalues the resulting dynamics are given by
\eq[final_decay]{
    \hode{t} = \exp\left(-\delta_t \diag\left(\decweight{i}\right) \right) \gruout{i}
}
where $\decweight{i} > 0$ is a parameter vector representing the negation of the eigenvalues.
We arrive at an exponential decay in-between observations, a type of dynamics used with GRU-updates in existing works \cite{decay_gru, brits}.
The positive restriction on $\decweight{i}$ ensures the stability of the dynamical system in the limit, which for the complete state means that $\latstate{t_j} \rightarrow \target{i}$ as $t_j \rightarrow \infty$.

On the other hand, if we instead allow $A$ to have complex eigenvalues we can decompose $A$ using the real Jordan form \cite{matrix_analysis}.
This makes $\Lambda$ block-diagonal with $2 \times 2$ blocks 
\eq[jordan_blocks]{
    C_j = \left[
    \begin{array}{cc}
         a_j & -b_j \\
         b_j & a_j
    \end{array}
    \right] 
}
corresponding to complex conjugate eigenvalues $a_j \pm b_j i$.
If we compute the matrix exponential for this $\Lambda$ we end up with a combination of exponential decay and periodic dynamics \cite{dynamical_systems}.
The solution gives dynamics that couple each pair of dimensions as
\als[final_periodic]{
    \left[\hode{t}\right]_{j:j+1} &=
    \exp\left(\delta_t a_j \right)\times\\
    &\left[
    \begin{array}{cc}
         \cos(b_j \delta_t) & -\sin(b_j \delta_t) \\
         \sin(b_j \delta_t) & \cos(b_j \delta_t)
    \end{array}
    \right]
    \left[\gruout{i}\right]_{j:j+1}.
}
We parametrize also these dynamics with $\decweight{i} = [-a_1, -a_3, \dots, -a_{\dh-1}, b_1, b_3, \dots, b_{\dh-1}]^\transpose > 0$.
Note that when $b_j \rightarrow 0$ these periodic dynamics reduce to the exponential decay in \refeq{final_decay}, but with the parameter for $-a_j$ shared across pairs of dimensions.
We consider both the exponential decay and the more general periodic dynamics as options for our model.
The periodic dynamics can naturally be seen as advantageous for modeling periodic data, as we will explore empirically in \refsec{periodic_exp}.

\subsection{Incorporating Observations from the Graph}
\label{sec:gru}
When node $n$ is observed at time $t_i$ the observation is incorporated into the latent state by a \gls{GRU} cell, incurring an instantaneous jump in the latent dynamics to a new value $\target{i} + \gruout{i}$.
Inspired by the continuous-time LSTM of \citeauth{neural_hawkes}, we extend the GRU cell to output also the parameters $\decweight{i}$, which define the dynamics of $\hode{n}$ for the next time interval.

So far we have considered each node of the graph as separate entities, but in a graph-based system future observations of a node can depend on the history of the entire graph.
To capture this we let the state of each node depend on observations and states in its graph neighborhood.
This is achieved by introducing \glspl{GNN} \cite{mpnn, gnn_survey} in the GRU update \cite{graphgru}.
We replace matrix multiplications with \glspl{GNN}, taking inputs both from the node $n$ itself and from its neighborhood $\neigh{n} = \set{m | (m,n) \in E}$. 
The type of \gls{GNN} we use is a simple version of a message passing neural network \cite{mpnn}, defined as
\als[graphconv]{
    \gnn&\left(\latent, \neighset{\latent[m]}\right)\\ 
    &= W_1 \latent + \frac{1}{\setsize{\neigh{n}}} \smashoperator[r]{\sum_{m \in \neigh{n}}} \edgew{m}{n} W_2 \latent[m]
}
where the matrices $W_1, W_2$ are learnable parameters shared among all nodes and $\edgew{m}{n}$ is an edge weight associated with the edge $(m,n)$.
The use of edge weights allows for incorporating prior information about the strength of connections in the graph.
We can additionally stack multiple such \gls{GNN} layers, append fully-connected layers and include non-linear activation functions in between.
The inclusion of multiple \gls{GNN} layers makes the GRU update dependent on a larger graph neighborhood.

In a standard \gls{GRU} cell the input and previous state are first mapped to three new representations using matrices $U$ and $W$ \cite{gru}.
These representations are then used to compute the state update.
In our \gls{GNN}-based \gls{GRU} update the matrices are replaced by \glspl{GNN} and we require seven such intermediate representations to update $\gruout{i}$, $\target{i}$ and $\decweight{i}$, as shown below.
The node states are combined as
\eq[graphgruh]{
    [\gruparth{1}, \dots, \gruparth{7}]^\transpose = \gnn^U\left(\latstate{t_i}, \neighset{\latstate[m]{t_i}}\right)
}
where the resulting vector is split into seven equally sized chunks.
Another \gls{GNN} is then used for the combined observations and input features $\gruinput = [\obsvec, \featurevec]^\transpose$,
\eq[graphgrux]{
    [\grupartx{1}, \dots, \grupartx{7}]^\transpose =
    \gnn^W\left(\gruinput, \neighset{\b{\tilde{x}}_i^m}\right).
}
Note that while all nodes might not be observed at time $t_i$, $\b{\tilde{x}}_{i}^{m} = \b{0}$ for any unobserved $m$ and thus do not contribute to the sum over neighbors in the \gls{GNN}.
With the combined information from the graph neighborhood the full \gls{GRU} update is computed as
\newcounter{subeq}
\newcommand{\stags}{\addtocounter{equation}{+1}\setcounter{subeq}{0}}
\newcommand{\stag}{\addtocounter{subeq}{+1} \tag{\theequation\alph{subeq}}}
\newcommand{\grueqdist}{7pt}
\al[grur]{
    \stags
    \grur &= \gruline{1} \stag \\ 
    \label{eq:gruz}
    \gruz &= \gruline{2} \stag \\ 
    \label{eq:gruh}
    \gruh &= \tanh(\grupartx{3} + (\grur \odot \gruparth{3}) + \b{b}_{3}) \stag\\ 
    \label{eq:gruout}
    \target{i} + \gruout{i}  &= (\b{1}-\gruz) \odot \latstate{t_i} + \gruz \odot \gruh \stag \\[\grueqdist] 
    \label{eq:grurt}
    \stags
    \grurt &= \gruline{4} \stag \\ 
    \label{eq:gruzt}
    \gruzt &= \gruline{5} \stag\\ 
    \label{eq:gruht}
    \gruht &= \tanh(\grupartx{6} + (\grurt \odot \gruparth{6}) + \b{b}_{6}) \stag\\ 
    \label{eq:gruoutt}
    \target{i} &= (\b{1}-\gruzt) \odot \target{i-1} + \gruzt \odot \gruht \stag\\[\grueqdist]
    \label{eq:decweight}
    \decweight{i} &= \log(\b{1} + \exp(\grupartx{7} + \gruparth{7} + \b{b}_{7}))
}
where $\sigma$ is the sigmoid function and $\b{b}_1$--$\b{b}_7$ learnable bias parameters. 
In \refeq{gruoutt} we let $\target{k} = \target{k-1}$ if $n \notin \obsn_{k}$.
Eq.~\ref{eq:grur}--\ref{eq:gruout} correspond to one GRU update, using the decayed state $\latstate{t_i}$.
Eq.\ \ref{eq:grurt}--\ref{eq:gruoutt} define a separate GRU update, but for the decay target $\target{i}$.
Note that we get $\gruout{i}$, the initial value of $\hode{t}$, implicitly from the difference between \refeq{gruout} and \refeq{gruoutt}.
Finally \refeq{decweight} computes the parameters $\decweight{i}$ defining the dynamics of $\hode{t}$ up until the next observation of node $n$.
The parameters of the GRU cell are shared for all nodes in the graph.
To capture any node-specific properties we parametrize initial states $\latstate{0}$ separately for all nodes and learn these jointly with the rest of the model.

\subsection{Predictions}
\label{sec:predictions}
The time-continuous dynamics ensure that there is a well-defined latent state $\latstate{t}$ in each node at each time point $t$.
The value of the time series can then be predicted at any time by applying a mapping $g\colon \R^{\dh} \rightarrow \R^{\dy}$ from this latent state to the prediction $\pred{j}$.

The addition of \glspl{GNN} into the GRU update makes the latent state dynamics of each node dependent on historical observations in its neighborhood.
However, since the GRU updates happen only when a node is observed, information from observed neighbors might not be incorporated immediately in the latent state.
Consider three consecutive time points $t_i < t_{i+1} < t_{i+2}$ s.t.\ $n \in \obsn_i$, $n \notin \obsn_{i+1}$.
Then any observation $\b{y}_{i+1}^m$ for $m \in \obsn_{i+1} \cap \neigh{n}$ will not be taken into account by the model for the prediction $\pred{i+2}$, as that prediction is based on a latent state with only information from time $t_i$.
To remedy this we choose also the predictive model $g$ to contain one or more \gls{GNN} layers,
\als[gnnpred]{
    \pred{j} &= 
\gnn^g\left([\latstate{t_j}, \featurevec[j]]^\transpose, \neighset{[\latstate[m]{t_j}, \b{x}^m_j]^\transpose}\right).
}
This way $g$ takes the latent states and input features of the whole neighborhood into account for prediction.
We name the full proposed model \acrlong{TGNN4I} (\textbf{\acrshort{TGNN4I}}).

\subsection{Loss Function}
\label{sec:loss}
In order to make predictions for arbitrary future time points we introduce a suitable loss function based on the time-continuous nature of the model.
Let $\yar$ be the prediction for node $m$ at time $t_j$, based on observations of all nodes at times $t \leq t_i$.
Define also a time-continuous weighting function $w\colon \R^{+} \rightarrow \R^{+}$ and the set $\tau_{m,i} = \set{j:m \in \obsn_{j} \wedge i <j}$ containing the indices of all times after $t_i$ where node $m$ is observed.
We do not include predictions from the first $\ninit$ time steps in the loss, treating this as a short warm-up phase.
The loss function for one graph-structured time series is then
\eq[new_loss]{
    \futureloss{\ell} = \frac{1}{\Nobs{}} 
    \sum_{m \in V} 
    \sum_{i = \ninit + 1 }^{N_t} 
    \sum_{j \in \tau_{m,i}} 
    \frac{\ell\left(\yar , \b{y}_j^m \right) w(t_j - t_i)}{j - \ninit - 1}
}
where $\ell$ is any loss function for a single observation and $\Nobs = \sum_{i=\ninit+2}^{N_t} \setsize{\obsn_i}$ the total number of node observations.
We use $\lmse$ with \gls{MSE} as $\ell$, but the framework is fully compatible with other loss functions as well.
This includes general probabilistic predictions with a negative log-likelihood loss.
Dividing by $j - \ninit - 1$, the number of times observation $j$ has been predicted, guarantees that later observations are not given a higher total weight.

The weighting function $w$ allows for specifying which time-horizons that should be prioritized by the model.
This choice is highly application-dependent and should capture which predictions that are of interest when later deploying the model in some real-world setting.
If we care about all time horizons, but want to prioritize predictions close in time, a suitable choice could be $w(\Delta_t)=\exp\left(-\frac{\Delta_t}{\Omega}\right)$.
It might also be desirable to focus predictive capabilities on a specific $\Delta_t$.
If we want predictions around $\Delta_t = \mu$ to be prioritized we can for instance use a Gaussian kernel as
\eq[gauss_kernel]{
    \textstyle
    w(\Delta_t)=\exp\left(-\left(\frac{\Delta_t - \mu}{\Omega}\right)^2 \right).
}

A limitation of the proposed loss function is the quadratic scaling in the number of time steps, as predictions are made from all times to all future observations.
This especially requires large amounts of memory for nodes that are observed at many time steps.
However, for many sensible
choices of $w$ predictions far into the future have a close to $0$ impact on the loss.
In practice we can utilize this to approximate $\futureloss{\ell}$ by only making predictions $\nmax$ time steps into the future.
This approximation explicitly corresponds to setting $\tau_{m,i} = \set{j : m \in \obsn_{j} \wedge i < j \leq i + \nmax{} }$ and changing the denominator in \refeq{new_loss} to $\min(\nmax, j - \ninit - 1)$.
Alternatively, we can select a weight function with finite support which implies that many terms in \refeq{new_loss} will be exactly zero.
This does however require more bookkeeping than the aforementioned truncation method.

\section{EXPERIMENTS}
\label{sec:experiments}
The \itgnn model was implemented\footnote{Our code and datasets are available at \url{https://github.com/joeloskarsson/tgnn4i}.} using PyTorch and PyG \cite{ptg}. 
We evaluate the model on a number of different datasets.
See \refapp{dataset_details} and \ref{sec:experiment_details} for details on the pre-processing and experimental setups used.
As the loss function $\lmse$ captures errors throughout an entire time series we adopt this also as our evaluation metric.
Given that the loss weighting $w$ used for training accurately represents how we value predictions at different time horizons, it is natural to use the same choice for evaluation.
In our experiments we rescale each time series so that $t \in [0,1]$ and use $w(\Delta_t) = \exp\left(-\frac{\Delta_t}{0.04}\right)$.
By inspecting $w$ and the time steps in the data we also choose a suitable $\nmax = 10$.

\newcommand{\listdyn}[1]{(\textbf{#1})}
We consider three versions of our \itgnn model: 
\listdyn{static} with a constant latent state in-between observations ($\hode{t} = \gruout{i} ~\forall \,t{\in} \,]t_i, t_j]$),
\listdyn{exponential} with the exponential decay dynamics from \refeq{final_decay} and
\listdyn{periodic} with the combined decay and periodic dynamics from \refeq{final_periodic}.
In all our experiments the training time of a single model on an NVIDIA A100 GPU is less than an hour and for the smallest dataset (\la) not more than 20 minutes.

\subsection{Baselines}
\label{sec:baselines}
\newcommand{\listbaseline}[1]{(\textbf{#1})}
We compare \itgnn to multiple baseline models.
As a simple starting point we consider a model that always predicts the last observed value in each node for all future time points \listbaseline{\predprev}.
\citeauth{decay_gru} propose the \grubase model for irregular time series, which we extend with our parametrization of the exponential decay and include as a baseline.
\grubase does not use the graph structure explicitly, so there are two ways to adapt this model to our setting. 
We can view the entire graph-structured time series as one series with $(\setsize{V}\dy)$-dimensional vectors at each time step \listbaseline{\grujoint}.
Alternatively, we can view the time series in each node as independent \listbaseline{\grunode}, which is essentially the same as \itgnn with all edges in the graph removed.
Two \textbf{\transbase} baselines are also included, used in the same (joint) and (node) configurations.
In these models the irregular observations are handled through attention masks and the use of timestamps in the sinusoidal positional encodings.
We also compare against the \textbf{\lgode} model of \citeauth{lgode} using the code provided by the authors.
We follow their proposed training procedure, where the model encodes the first half of each time series and has to predict the second half.
When computing $\lmse$ using \lgode we encode all observations up to $t_i$ and decode from that time point in order to get each $\yar$.
More details on the baseline models are given in \refapp{baseline_details}. %
An attempt was also made to adapt the RAINDROP model of \citeauth{raindrop} to our forecasting setting.
We were however unable to get useful predictions without making major changes to the model and it is therefore not included here.

\subsection{Traffic Data}
\label{sec:traffic_exp}
We experiment on the \bay{} and \la{} datasets, containing traffic speed sensor data from the California highway system \cite{dcrnn_traffic, pems_system}.
To be able to control the degree of irregularity, we start from regularly sampled data and choose subsets of observations.
We use the versions of the datasets pre-processed by \citeauth{dcrnn_traffic}.
Each dataset is split up into time series of 288 observations (1 day).
\bay contains 180 such time series with 325 nodes and \la 119 time series with 207 nodes.
We include the time of day and the time since the node was last observed as input features $\featurevec$.
In order to introduce irregularity in the time steps we next subsample each time series by keeping only 72 of the 288 observations. These $N_t = 72$ observation times are the same for all nodes.
However, from these subsampled time series we furthermore sample subsets containing 25\%--100\% of all $N_t \times \setsize{V}$ individual node observations.
This results in irregular observation time points and a fraction of nodes observed at each time.
Our additional pre-processing prevents us from a direct comparison with \citeauth{dcrnn_traffic}, as their method does not handle irregular observations.

\begin{table*}[t]
\centering
\caption{
Test $\lmse$ (multiplied by $10^2$) for the traffic datasets with different fractions of node observations. 
Where applicable we report mean \textpm{} one standard deviation across 5 runs with different random seeds.
The lowest mean $\lmse$ for each dataset and observation percentage is marked in bold.
}
\robustify\bfseries
\label{tab:traffic_res}

\begin{tabular}{@{}l
S[table-format=2.2(3),mode=text]
S[table-format=2.2(3),mode=text]
S[table-format=2.2(3),mode=text]
S[table-format=2.2(3),mode=text]
@{}}
\toprule
                  & \multicolumn{4}{c}{\textbf{\bay}}\\ 
\cmidrule(lr){2-5}
\textbf{Model}    & \multicolumn{1}{c}{25\%}  & \multicolumn{1}{c}{50\%}  & \multicolumn{1}{c}{75\%}  & \multicolumn{1}{c}{100\%}\\ \midrule
Predict Previous  & 26.32          & 18.60          & 15.25          & 13.50          \\
\grujoint         & 18.79 \pm 0.07 & 18.27 \pm 0.10 & 17.93 \pm 0.08 & 17.75 \pm 0.12 \\
\grunode          & 8.79\pm0.06  & 6.62\pm0.02  & 5.82\pm0.06  & 5.49\pm0.06  \\
\transjoint       & 12.05\pm1.19 & 13.13\pm2.59 & 12.21\pm1.95 & 11.09\pm1.38 \\
\transnode        & 16.49\pm0.17 & 14.44\pm0.48 & 13.20\pm0.56 & 13.16\pm1.23 \\
\lgode            & 27.00          & 24.93          & 24.71          & 23.52          \\
\itgnn (static)   & 7.41\pm0.09  & 5.98\pm0.07  & 5.29\pm0.08  & 4.89\pm0.05  \\
\itgnn (exponential)             & \bfseries 7.10\pm0.07 & \bfseries 5.78\pm0.05 & 5.23\pm0.03          & \bfseries 4.87\pm0.09 \\
\itgnn (periodic)                & \bfseries 7.10\pm0.09 & 5.80\pm0.08          & \bfseries 5.22\pm0.09 & \bfseries 4.87\pm0.02 \\ 
\midrule
                  & \multicolumn{4}{c}{\textbf{\la}}\\ 
\cmidrule(lr){2-5}
\textbf{Model}    & \multicolumn{1}{c}{25\%}  & \multicolumn{1}{c}{50\%}  & \multicolumn{1}{c}{75\%}  & \multicolumn{1}{c}{100\%}\\ \midrule
Predict Previous  & 9.86           & 7.54           & 6.52           & 6.04           \\
\grujoint         & 8.38\pm0.05  & 8.03\pm0.04  & 7.89\pm0.03  & 7.80\pm0.02  \\
\grunode          & 4.36\pm0.08  & 3.62\pm0.07  & 3.28\pm0.08  & 3.16\pm0.04  \\
\transjoint       & 5.70\pm1.41  & 7.17\pm1.66  & 5.95\pm1.90  & 6.11\pm1.80  \\
\transnode        & 7.01\pm0.31  & 6.34\pm0.24  & 5.84\pm0.23  & 5.96\pm0.50  \\
\lgode            & 8.51           & 7.35           & 6.71           & 6.24           \\
\itgnn (static)   & 3.86\pm0.02  & 3.31\pm0.02  & 3.03\pm0.02  & 2.88\pm0.02  \\
\itgnn (exponential)             & \bfseries 3.68\pm0.05 & \bfseries 3.18\pm0.03 & \bfseries 2.97\pm0.03 & \bfseries 2.86\pm0.04 \\
\itgnn (periodic) & 3.69\pm0.02  & 3.19\pm0.04  & 3.01\pm0.05  & 2.88\pm0.03  \\ \bottomrule
\end{tabular}

\end{table*}
\begin{figure}[tb]
    \centering
    \includegraphics[width=\columnwidth]{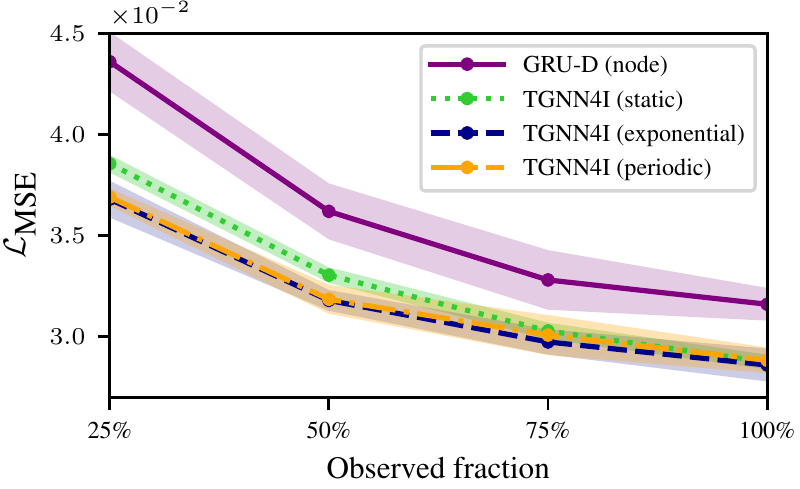}
    \caption{
    Test $\lmse$ on \la traffic data for the best performing models from \reftab{traffic_res}. 
    Shaded areas correspond to 95\% confidence intervals based on re-training models with 5 random seeds.
    }
    \label{fig:la_res_plot}
\end{figure}

We report results for both datasets in \reftab{traffic_res} and highlight the best performing models on \la in \reffig{la_res_plot}.
\grujoint has a hard time modeling all nodes jointly, often not performing better than the simple \predprev baseline.
The \transbase models achieve somewhat better results, but still not competitive with \itgnn.
We additionally note that the Transformers can be highly sensitive to the random seed used for initialization, something that we have not observed for other models. 
Comparing \itgnn and \grunode in \reffig{la_res_plot} it can be noted that the importance of using the graph structure increases when there are fewer observations.
Out of the different versions of \itgnn the exponential and periodic dynamics show a clear advantage over the static one, with the largest difference for the most sparsely observed data.
We have observed that the periodic models output only low frequencies, resulting in dynamics and results similar to the model with exponential decay.
While the periodic dynamics in \refeq{final_periodic} have fewer degrees of freedom when reduced to pure exponential decay, this does not seem to hurt the performance in this example.

We found that the training time of \lgode scales poorly to large graphs, limiting us to only training a single model for each dataset.
The predictions are however quite poor, especially on the \bay data.
While the model seems to learn something more than just predicting the mean, it is not competitive with our \itgnn model.
We believe that the poor performance of \lgode can be explained by a combination of multiple things:
\begin{inparaenum}[(1)]
    \item The \lgode model is primarily designed for data with clear continuous underlying dynamics, which might not match this type of traffic data.
    \item When the model is trained as proposed by \citeauth{lgode}, it can require large amounts of data.
    For some of the experiments in the original paper 20\,000 sequences are used for training, while we use less than 150.
    \item The slow training has limited possibilities for exhaustive hyperparameter tuning on our datasets.
    Training one \lgode model on the \bay data takes us over 50 hours.
\end{inparaenum}

\subsection{USHCN Climate Data}
\label{sec:climate_exp}
Irregular and missing observations are common problems in climate data \cite{analysis_of_climate_data}.
The \gls{USHCN} daily dataset contains over 50 years of measurements of multiple climate variables from sensor stations in the United States \cite{ushcn}.
We use the pre-processing of \citeauth{gru_ode_bayes} to clean and subsample the data.
The target variables chosen are minimum and maximum daily temperature ($\tmin$ and $\tmax$), which we model as separate datasets.
While existing works \cite{gru_ode_bayes, cru} have treated time series from different sensor stations as independent, we model also the spatial correlation by constructing a 10-nearest-neighbor graph using the sensor positions.
Each full dataset contains 186 time series of length $N_t = 100$ on a graph of 1123 nodes.
The pre-processed USHCN data is sparsely observed with only around 5\% of potential node observations present.

\begin{table}[tb]
\centering
\caption{
Test $\lmse$ (multiplied by $10^2$) for the two USHCN climate datasets. 
}
\label{tab:ushcn_res}
\robustify\bfseries
\begin{tabular}{@{}lS[table-format=2.2(3),mode=text]S[table-format=2.2(3),mode=text]@{}}
\toprule
                   & {$T_{\min}$} & $T_{\max}$ \\ 
\midrule
\predprev   & 16.88                              & 17.18                       \\
\grujoint   & 8.03\pm0.23                       & 7.97\pm0.19                       \\
\grunode    & 13.12\pm0.03                       & 13.67\pm0.04                       \\
\transjoint    & 7.36\pm0.41                       & 7.37\pm0.28                       \\
\transnode    & 15.68\pm0.32                       & 15.74\pm0.34                       \\
\itgnn (static)    & 6.97\pm0.05                       & 6.86\pm0.04                       \\
\itgnn (exponential)     & \bfseries 6.72 \pm 0.04                       & \bfseries 6.60\pm0.04      \\ 
\itgnn (periodic)     & \bfseries 6.72\pm0.05                       & 6.63\pm0.03                       \\ \bottomrule
\end{tabular}
\end{table}

We report results on both datasets in \reftab{ushcn_res}.
Due to the large size of the graph it was not feasible to apply the \lgode model here.
We note that for these datasets the (joint) baselines clearly outperform the (node) versions.
For \grubase this is the opposite of what we saw in the traffic data.
This can be explained by the fact that climate data has strong spatial dependencies.
The (joint) models can to some extent learn to pick up on these, while for (node) no information can flow between nodes.
The best results are however achieved by \itgnn, showing the added benefit of utilizing the spatial graph.

\subsection{Synthetic Periodic Data}
\label{sec:periodic_exp}
In the previous experiments, using periodic dynamics with \itgnn has not added any value. 
Instead, the learned dynamics have been largely similar to just using exponential decay.
This should to some extent be expected, as none of the previous datasets show any clear periodic patterns at the considered time scales.
To investigate the possible benefits of the periodic dynamics we instead create a synthetic dataset with periodic signals propagating over a graph.

The synthetic dataset is based on a randomly sampled directed acyclic graph with 20 nodes.
We define a periodic base signal 
\eq[base_signal]{
    \simbase = \sin(\phi^n t + \eta^n)
}
with random parameters $\phi^n$ and $\eta^n$ for each node $n$. 
The target signal $\simfull$ in each node is then defined through
\begin{subequations}
\label{eq:synth_signal}
\al[synth_node_signal]{
    \simsignal{t} &= \simbase + \frac{0.5}{\setsize{\neigh{n}}} \sum_{m \in \neigh{n}} \simsignal[m]{t - 0.05}\\
    \simfull &= \simsignal{t} + \simnoise
}
\end{subequations}
where $\simnoise$ is Gaussian white noise with standard deviation 0.01.
The target signal in each node depends on the base signal in the node itself and the signals in neighboring nodes at a time lag of 0.05.
To construct one time series we sample $\simfull$ at 70 irregular time points on $[0,1]$.
In total we sample 200 such time series and keep 50\% of the node observations in each. 

\begin{table}[tb]
\centering
\caption{
Test $\lmse$ (multiplied by $10^2$) for synthetic data. 
}
\label{tab:periodic_res}
\robustify\bfseries
\begin{tabular}{@{}l@{\hspace{\tabcolsep}}S[table-format=2.2(3),mode=text]S[table-format=2.2(3),mode=text]S[table-format=2.2(3),mode=text]@{}}
\toprule
&\multicolumn{1}{c}{\textbf{Static}}& \multicolumn{1}{c}{\textbf{Exponential}} & \multicolumn{1}{c}{\textbf{Periodic}}\\ \midrule
\grunode & 8.88\pm0.36       & 3.13\pm0.06                         & 2.81\pm0.04                       \\
\itgnn     & 15.12\pm0.05       & 2.91\pm0.17                         & \bfseries 1.95\pm0.11              \\ 
\midrule
Predict Prev.     &        & 27.52                         &               \\
\multicolumn{2}{@{}l}{\transjoint}     & 23.19\pm0.38                         &               \\
\multicolumn{2}{@{}l}{\transnode}     & 15.39\pm0.05                         &               \\ 
\lgode     &        & 16.61\pm0.23                         &               \\ 
\bottomrule
\end{tabular}
\end{table}
We train versions of the \grunode and \itgnn models with different latent dynamics on the synthetic data.
Also our other baselines are included for comparison.
Results are reported in \reftab{periodic_res}.
For both \grunode and \itgnn we see a large difference between the different types of latent dynamics.
The periodic dynamics seem to help the model to keep track of the base signal in the node and its neighborhood in order to achieve accurate future predictions.
While this is a synthetic example, periodic behavior is prevalent in much time series data and being able to explicitly model this in the latent state can be highly advantageous. 
Attempts were made to also train the \grujoint model on this dataset, but it failed to pick up on any patterns and ended up only predicting a constant value for all nodes and times.

\subsection{Loss Weighting}
\label{sec:loss_exp}

\begin{figure}[tb]
\centering
    \includegraphics[width=\columnwidth]{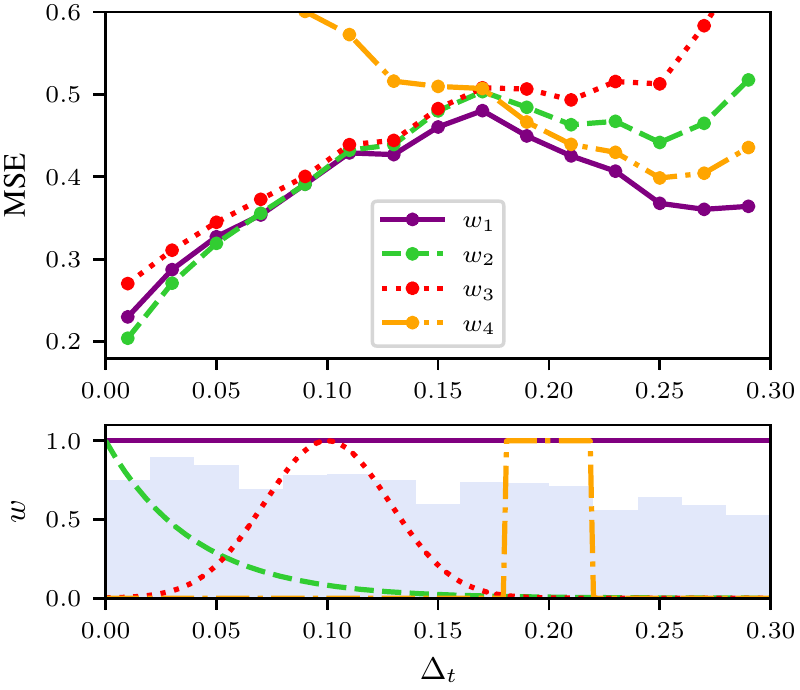}
    \caption{
    \gls{MSE} (Top) for predictions at time $\Delta_t$ in the future for models trained with different loss weighting $w$ (Bottom).
    The weighting functions are described in \refeq{w_exp}--\ref{eq:w_exp_end}.
    To compute the MSE all predictions in the test set were binned based on their $\Delta_t$, with bin width 0.02.
    The bottom subplot also shows a histogram of the number of predictions in each bin.
    }
    \label{fig:loss_weighting}
\end{figure}

To investigate the impact of the loss weighting function $w$ we trained four \itgnn models on the subsampled \bay dataset with 25\% observations.
We used exponential dynamics and considered the weighting functions\begin{subequations}
\al[w_exp]{ %
    w_1(\Delta_t) &= 1\\
    w_2(\Delta_t) &= \exp\left(-\frac{\Delta_t}{0.04}\right)\\
    w_3(\Delta_t) &= \exp\left(-\left(\frac{\Delta_t - 0.1}{0.02}\right)^2 \right)\\
    \label{eq:w_exp_end}
    w_4(\Delta_t) &= \indicator{\Delta_t \in [0.18, 0.22]}.
}
\end{subequations}
\reffig{loss_weighting} shows the test \gls{MSE} for the trained models at different $\Delta_t$ in the future, as well as plots of the weighting functions.
As an example, the prediction $\yar$ has $\Delta_t = t_j - t_i$.

We note that the choice of $w$ can have substantial impact on the error of the model at different time horizons.
The exponential weighting in $w_2$ makes the model focus heavily on short-term predictions.
This results in better predictions for low $\Delta_t$.
At the shortest time horizon the exponential weighting yields an 11\% improvement over the model trained with constant $w_1$, but this comes at the cost of far higher errors for long-term predictions.
Interestingly the $w_1$ model gives better predictions at all time horizons than the models with $w_3$ and $w_4$, which focus on predictions at some specific time ahead.
We believe that there can be a feedback effect benefiting the constant weighting, where learning to make good short-term predictions also aid the learning of long-term prediction, for example by finding useful intermediate representations.
A drawback of weighting with $w_1$ is however that since the loss never approaches 0 there is no properly motivated choice of $\nmax$ for our loss approximation.
As the model trained with $w_4$ still gives good predictions in the interval $[0.18, 0.22]$, there can still be practical reasons to choose such a weighting. 
With this choice we could reduce the innermost sum in \refeq{new_loss} to only those $j$:s that lie in the interval of interest.

\section{DISCUSSION}
We have proposed a temporal \gls{GNN} model that can handle both irregular time steps and partially observed graphs.
By defining latent states in continuous time our model can make predictions for arbitrary time points in the future.
In this section we discuss some details and limitations of the approach, and also give some pointers to interesting directions for future work.

\subsection{Efficient Implementation}
In order to efficiently implement the training and inference of \itgnn there is a key design choice between
\begin{inparaenum}[(1)]
\item storing everything in dense matrices and utilizing massively parallel GPU-computations, and
\item utilizing sparse representations in order to avoid computing values that will never be used.
\end{inparaenum}
Which of these is to be preferred depends on the sparsity of observations in the data.
Our implementation follows the massively parallel approach, using binary masks for keeping track of $\obsn_i$ at each time point.
In order to scale \itgnn to massive graphs in real world scenarios it could be interesting to consider a version of the model distributed over multiple machines, perhaps directly connected to sensors producing the input data.
A sparse implementation would be strongly preferred for such an extension.

\subsection{Linear and Neural ODEs}
The dynamics of \itgnn are defined by a linear \gls{ODE} and additionally restricted by the assumption of unique eigenvalues in $A$.
This has the benefit of a closed form solution that is efficient to compute, but also limits the types of dynamics that can be learned.
Our approach can be contrasted with Neural \glspl{ODE} \cite{neural_ode}, that allow for learning more expressive dynamics.
Neural \glspl{ODE} do however lack closed form solutions and require using numerical solvers \cite{on_neural_odes}.
This incurs a trade-off between speed and numerical accuracy.
In experiments we have compared \itgnn with the \lgode model \cite{lgode}, which uses a Neural \gls{ODE} decoder.
The slow training of the \lgode model can to a large extent be attributed to the numerical \gls{ODE} solver.
While more complex latent dynamics can be useful for some datasets, it can also be argued that simpler dynamics can be compensated with a high enough latent dimension $d_h$ and a flexible enough predictive model $g$ \cite{cru}.

\subsection{Societal and Sustainability Impact}
While our contributions are purely methodological, many applications of graph-based and spatio-temporal data analysis have a clear societal impact.
Our example applications of traffic and climate modeling both have potential to aid efforts of transforming society in more sustainable directions, such as those described in the United Nations sustainable development goals 11 and 13 \cite{tackling_climate_ml, un_sdgs}.
Traffic modeling allows us to both understand travel behavior and predict future demand.
This can enable optimizations of transport systems, both improving the experience of travelers and reducing the environmental impact.
Integrating machine learning methods with climate modeling has potential to speed up simulations and increase our understanding of the climate around us.
However, it is surely also possible to find applications of our method with a damaging impact on society, for example through undesired mass-surveillance.

\subsection{Future Work}
We consider a setting where the graph structure is both known and constant over time.
In some practical applications it is not obvious how to construct the graph describing the system.
To tackle this problem our model could be combined with approaches for also learning the graph structure \cite{data_analytics_on_graphs3, raindrop}. 
Extending our method to dynamic graphs, that evolve over time, would not require any major changes and could be an interesting direction for future experiments.

Our focus has been on forecasting, but the model could also be trained for other tasks.
The time-continuous latent state in each node could be used for imputing missing observations or performing sequence segmentation.
Also classification tasks are possible, either classifying each node separately or the entire graph-structured time series.

While our model can produce predictions at arbitrary time points, an extension would be to also predict the time until the next observation occurs.
One way to achieve this would be to let the latent state parametrize the intensity of a point process \cite{neural_hawkes, jump_sde}.
Building such point process models on graphs could be an interesting future application of our model.

\subsubsection*{Acknowledgments}
This research is financially supported by the Swedish Research Council via the project
\emph{Handling Uncertainty in Machine Learning Systems} (contract number: 2020-04122),
the Excellence Center at Linköping--Lund in Information Technology (ELLIIT),
and
the Wallenberg AI, Autonomous Systems and Software Program (WASP) funded by the Knut and Alice Wallenberg Foundation.
The computations were enabled by the Berzelius resource at the National Supercomputer Centre, provided by the Knut and Alice Wallenberg Foundation.
All datasets available online were accessed from the Linköping University network.

\bibliography{references}

\appendix
\onecolumn

\section{TABLE OF NOTATION}
\label{sec:notation}
The notation used is listed in \reftab{notation}.
\newcommand{\notationrowfirst}[4]{$#1$ & #2 & \multirow{#4}{*}{#3}\\}
\newcommand{\notationrowsec}[4]{\notationrowfirst{#1}{#2}{Section \ref{sec:#3}}{#4}}
\newcommand{\notationrow}[2]{$#1$ & #2 & \\}
\begin{table}[thp]
\centering
\caption{
List of notation 
}
\label{tab:notation}
\begin{tabular}{@{}cll@{}}
\toprule
\textbf{Notation} & \textbf{Description} & \textbf{Defined in}\\
\midrule
\notationrowfirst{\diag(\b{v})}{Diagonal matrix with entries of vector $\b{v}$ on the diagonal}{}{3}
\notationrow{\left[\b{v}\right]_{j:j+1}}{Entries $j$ and $j+1$ of vector $\b{v}$}
\notationrow{\indicator{\cdot}}{Indicator function}
\midrule

\notationrowsec{\G}{The underlying graph of the time series}{setting}{9}
\notationrow{V, E}{Node and edge set of $\G$}
\notationrow{t_i}{Time point where at least one node is observed}
\notationrow{N_t}{Number of time points in one graph-structured time series}
\notationrow{\obsn_i}{Set of nodes observed at $t_i$}
\notationrow{\obsvec}{Observed value in node $n$ at time point $t_i$. Equal to $\b{0}$ if node $n$ is not observed at time $t_i$.}
\notationrow{\featurevec}{Features of node $n$ at time point $t_i$. Equal to $\b{0}$ if node $n$ is not observed at time $t_i$.}
\notationrow{\dx, \dy, \dh}{Dimensionality of $\featurevec$, $\obsvec$ and $\latstate{t}$}
\notationrow{\latstate{t}}{Time-continuous latent state of node $n$}
\midrule

\notationrowsec{\target{i}}{Static part of $\latstate{t}$ after time $t_i$}{dynamics}{10}
\notationrow{\hode{t}}{Dynamic part of $\latstate{t}$ defined by a linear \gls{ODE}}
\notationrow{A}{Coefficient matrix in the \gls{ODE} defining $\hode{t}$}
\notationrow{\gruout{i}}{Initial value of $\hode{t_i}$ in the \gls{ODE} from time $t_i$ onward}
\notationrow{\delta_t}{Elapsed time since $t_i$}
\notationrow{Q}{Matrix containing the eigenvectors of $A$}
\notationrow{\Lambda}{Diagonal matrix containing the eigenvalues of $A$}
\notationrow{\decweight{i}}{Parameters defining the dynamics of $\hode{t}$ after $t_i$}
\notationrow{C_j}{$2\times2$ block matrix in real Jordan form of $A$}
\notationrow{a_j \pm b_j i}{Complex conjugate eigenvalue pair of $A$}
\midrule

\notationrowsec{\neigh{n}}{Neighbors (parents) of node $n$}{gru}{12}
\notationrow{W_1, W_2}{Matrices used in \gls{GNN}}
\notationrow{e_{m,n}}{Edge weight associated with the edge $(m,n)$}
\notationrow{\gnn^U}{\gls{GNN} combining latent states in neighborhood of node $n$}
\notationrow{\gruparth{1}$ -- $\gruparth{7}}{Output of $\gnn^U$ for node $n$ at time $t_i$. Intermediate representations used in \gls{GRU} update.}
\notationrow{\gruinput}{Concatenation of $\obsvec$ and $\featurevec$}
\notationrow{\gnn^W}{\gls{GNN} combining features and observations in neighborhood of node $n$}
\notationrow{\grupartx{1}$ -- $\grupartx{7}}{Output of $\gnn^W$ for node $n$ at time $t_i$. Intermediate representations used in \gls{GRU} update.}
\notationrow{\grur, \gruz, \gruh}{Intermediate representations in \gls{GRU} update for $\latstate{t}$}
\notationrow{\sigma}{Sigmoid function}
\notationrow{\b{b}_1$--$\b{b}_7}{Bias parameters in \gls{GRU} cell}
\notationrow{\grurt, \gruzt, \gruht}{Intermediate representations in \gls{GRU} update for $\target{i}$}
\midrule

\notationrowsec{g}{Predictive model}{predictions}{3}
\notationrow{\pred{j}}{Predicted value of node $n$ at time $t_j$}
\notationrow{\gnn^g}{\gls{GNN} used as predictive model $g$}
\midrule

\notationrowsec{\yar}{Predicted value of node $m$ at time $t_j$ based on observations before or at time $t_i$}{loss}{10}
\notationrow{w}{Loss weighting function}
\notationrow{\tau_{m,i}}{Set of indices of all times after $t_i$ where node $m$ is observed}
\notationrow{\ninit}{Length of warm-up phase where predictions are not included in loss}
\notationrow{\futureloss{\ell}}{Loss function for one whole graph-structured time-series}
\notationrow{\ell}{General loss function for a single observation}
\notationrow{\lmse}{$\futureloss{\ell}$ with \acrlong{MSE} as loss for each observation}
\notationrow{\Nobs}{Number of node observations in time-series}
\notationrow{\Delta_t}{Time difference between last observation and prediction time}
\notationrow{\nmax}{Maximum number of time steps to predict ahead in approximation of $\futureloss{\ell}$}

\bottomrule
\end{tabular}
\end{table}

\section{IMPLEMENTATION AND TRAINING DETAILS FOR \itgnn}
\label{sec:implementation_details}
The presentation of \itgnn in Section \ref{sec:method} describes how the model processes a single graph-structured time-series.
In practice we use batches of multiple such time series during training and inference.
When working on batches there is an additional sum in the definition of $\futureloss{\ell}$, computing the mean over all samples in the batch.
While we assume that all time series in a batch share the same $N_t$, the exact time points $\set{t_i}_{i=1}^{N_t}$ can differ.
In all experiments we use $\ninit = 5$ and minimize the loss using the Adam optimizer \citeapp{adam}.

Both $\gnn^U$ and $\gnn^W$ contain only \gls{GNN} layers, while the predictive model $g$ contains a sequence of \gls{GNN} layers followed by a sequence of fully connected layers.
We use ReLU activation functions in between all layers.

The input $\gruinput[m]$ to $\gnn^W$ is $\b{0}$ for neighbors that are not observed at time $t_i$.
However, there could be an observation of neighbor $m$ where the observed value and input features are exactly $\b{0}$.
To help the model differentiate between these two cases we additionally include an indicator variable $\indicator{m \in \obsn_i}$ in $\gruinput[m]$.

\section{BASELINE MODELS}
\label{sec:baseline_details}
\subsection{\predprev}
The \predprev baseline does not require any training.
Predictions are computed as $\yar = \b{y}^m_{\lastobsi(i)} ~\forall j$, where $\lastobsi(i) = \max \set{k : k \leq i \wedge m \in \obsn_k}$.
If there is no earlier observation of node $m$ the predictions is just $\b{0}$.

\subsection{GRU-D (node)}
GRU-D (node) is essentially a version of \itgnn without the \gls{GNN} components.
The \glspl{GNN} $\gnn^U$ and $\gnn^W$ in the \gls{GRU} update are replaced by matrix multiplications and the predictive model $g$ includes only fully connected layers.
Because of this no information can flow between nodes, and time-series in different nodes are treated as independent.
The \grunode model uses the exponential dynamics for the latent state.
To stay consistent with the other models we still process all nodes in the graph concurrently.
This means that a batch size of $B$ for \grunode means that we process $B \times N$ independent node time-series.

\subsection{GRU-D (joint)}
The \grujoint model is defined similar to \grunode, but modeling all nodes jointly.
All node observations are concatenated into one long vector $\b{y}_i = \left[\b{y}_i^1, \dots, \b{y}_i^{\setsize{V}}\right]^\transpose$ and similarly the features are concatenated as $\b{x}_i = \left[\b{x}_i^1, \dots, \b{x}_i^{\setsize{V}}\right]^\transpose$.
This time series is then modeled using a single latent state, also based on the exponential dynamics.
In \grujoint there is however a \gls{GRU} update at each $t_i$, as at least one of the nodes is observed at each time point.
We can think of this setup as a graph with a single node, for which the observations are high-dimensional.
The high-dimensional predictions are split up and re-assigned to the original nodes in the graph for computing the loss.
Note also that many entries in each $\b{y}_i$ and $\b{x}_i$ will be zero, corresponding to the nodes that are not observed.
We again include indicators $\indicator{m \in \obsn_i}$ as input to the \gls{GRU}, but here for all nodes.

\subsection{\lgode}
For the \lgode model we use the code provided by the authors \cite{lgode}\footnote{\url{https://github.com/ZijieH/LG-ODE}}, making only small modifications to the data loading in order to correctly handle our datasets.
The original code does not use a validation set, instead evaluating the model on the test set after each training epoch.
We change this step to use validation data and save the model from the epoch with the lowest validation error.
That model is then loaded and evaluated on the test data.
Due to the high training time we have not been able to perform exhaustive hyperparameter tuning for the \lgode model.
We have mainly used the default parameters of \citeauth{lgode} or a slightly smaller version of the model.
The smaller version has halved dimensions for hidden layers in the \gls{GNN} (from 128 to 64) and augmentation (from 64 to 32) used in the \gls{ODE} decoder.

For computing $\lmse$ using the \lgode model we need to compute each necessary $\yar$.
This is done by encoding all observations up to time $t_i$ and then decoding $\nmax$ time steps into the future.
It should be noted that the model is still trained as proposed by \citeauth{lgode}, by encoding the first half of each time series and predicting the second half.
During training each encoded time series is $N_t / 2$ long, but when computing $\lmse$ on the test set the length of the sequence being encoded varies.
We have however not noticed any higher errors for time steps with a shorter or longer encoded sequence than that used during training.

\subsection{Transformers}
The \transbase \cite{attention_all_you_need} baselines use an encoder-decoder approach similar to \lgode.
During training a prediction time $t_i$ is however randomly sampled as $i \sim \uniform{\set{\ninit, \dots, N_t - \nmax}}$.
Each sequence is then encoded up to time $t_i$ and decoded over the next $\nmax$ time steps to produce predictions.
The $\lmse$ loss is used also for the \transbase models, but only based on this one prediction per time series.

The irregular time steps are handled by sinusoidal encodings concatenated to the input of both the encoder and decoder.
Instead of basing these on the sequence index $i$, the exact timestamp $t_i$ is used in
\eq[encoding_angle]{
    \sinangle{i} = \frac{t_i}{0.1^{2i/\dh}}
}
to then compute the full encoding vector $\left[
    \sin(\sinangle{1}), \dots, \sin(\sinangle{\lfloor\dh/2\rfloor}), 
    \cos(\sinangle{1}), \dots, \cos(\sinangle{\lfloor\dh/2\rfloor})
    \right]^\transpose$.

Unobserved nodes are in \transnode handled by masking the attention mechanism.
For each node $n$, this prevents the model from attending to encoded time steps $j$ s.t.\ $n \notin \obsn_{j}$.
In \transjoint the same approach as in \grujoint is instead used, where indicator variables are included as input.

The hyperparameters defining the \transbase architectures differ somewhat from the other models.
We still let $\dh$ represent the dimensionality of hidden representations, but here also tune the number of transformer layers stacked together.

\section{DATASETS}
\label{sec:dataset_details}
\subsection{Traffic Data}
For the \bay{} and \la{} datasets we use the versions pre-processed by \citeauth{dcrnn_traffic}, where weighted graphs are created based on thresholded road-network distances.
We additionally remove edges from any node to itself and drop nodes not connected to the rest of the graph.
The original time series is then split into sequences of length 288, corresponding to one day of observations at 5 minute intervals.
In each such sequence we uniformly sample only $N_t = 72$ time points to keep, introducing irregularity between the time steps.
Next we create the set $\set{(n, i)}_{n \in V, i=1, \dots, N_t}$ with indices of all single node observations.
We uniformly sample a fraction of this set as the observations to keep, independently for each sequence.
The percentages in Table \ref{tab:traffic_res} refer to the percentage of observations kept from this set.
This step introduces further irregularity as all nodes will generally not be observed at each time step.
The \la dataset has some observations missing initially, meaning that we can never get to exactly 100\% observations for that dataset.
From this pre-processing we end up with 180 time series in the \bay dataset and 119 time series in the \la dataset.
We randomly assign 70\% of these to the training set, 10\% to the validation set and 20\% to the test set.

\subsection{USHCN Climate Data}
The \gls{USHCN} daily data is openly available\footnote{\url{https://cdiac.ess-dive.lbl.gov/ftp/ushcn_daily/}} together with the positions of all sensor stations.
We use the pre-processing of \citeauth{gru_ode_bayes} to clean and subsample the data\footnote{A script for pre-processing the USHCN data is available together with their code at \url{https://github.com/edebrouwer/gru_ode_bayes}.}.
We choose the longer version of the time series, with observations between the years 1950 and 2000, but split this into multiple sequences of 100 days.

In order to build the spatial graph we perform an equirectangular map projection of the sensor station coordinates and then connect each station to its 10 nearest neighbors based on euclidean distance.
While there are many options for how to create this type of spatial graph, we have found the 10-nearest-neighbor approach to work sufficiently well in practice.
Further investigating different methods for building spatial graphs is outside the scope of this paper.
We additionally add edge weights to this graph following a similar method as \citeauth{dcrnn_traffic} did for the traffic data.
For sensor stations $m$ and $n$ at a distance $d_{m,n}$ we assign the edge $(m,n)$ the weight 
\eq[ushcn_edge_weight]{
e_{m,n} = \exp\left(-\left(\frac{d_{m,n}}{4 \sigma_e}\right)^2\right)
}
where $\sigma_e$ is the standard deviation of all distances associated with edges.
The constant 4 is chosen such that the furthest distance gets a weight close to 0 and the nearest distance a weight close to 1.
For the \gls{USHCN} data the only input feature is elapsed time since the node was last observed.
We use the same 70\%/10\%/20\% training/validation/test split as for the traffic data.

We build two datasets from the \gls{USHCN} data, one with the daily minimum temperature $\tmin$ as target variable and one with the daily maximum temperature $\tmax$.
The reason for separating these target variables, instead of using $d_y = 2$, is that the pre-processed data contains time points where only one of these is observed.
Our model is not designed for such a setting where we do not observe all dimensions of $\obsvec$.
Extending \itgnn to handle this is left as future work.
The \gls{USHCN} data also contains other climate variables, related to precipitation and snow coverage.
These time series are less interesting to directly evaluate our model on, as many entries are just 0.
Properly modeling these would require designing a suitable likelihood function and taking into account that some sensor stations never get any snow.
As this would shift the focus from the core problem we choose to restrict our attention to $\tmin$ and $\tmax$.

\subsection{Synthetic Periodic Data}
To create the graph for the synthetic data we first sample 20 node positions uniformly over $[0,1]^2$.
We then create an undirected graph using a Delaunay triangulation \citeapp{triangulations_book} based on these positions.
This undirected graph is turned into a directed acyclic graph by choosing a random ordering of the nodes and removing edges going from nodes later in the ordering to those earlier.

\begin{figure*}[tbh]
\begin{center}
\includegraphics[width=\linewidth]{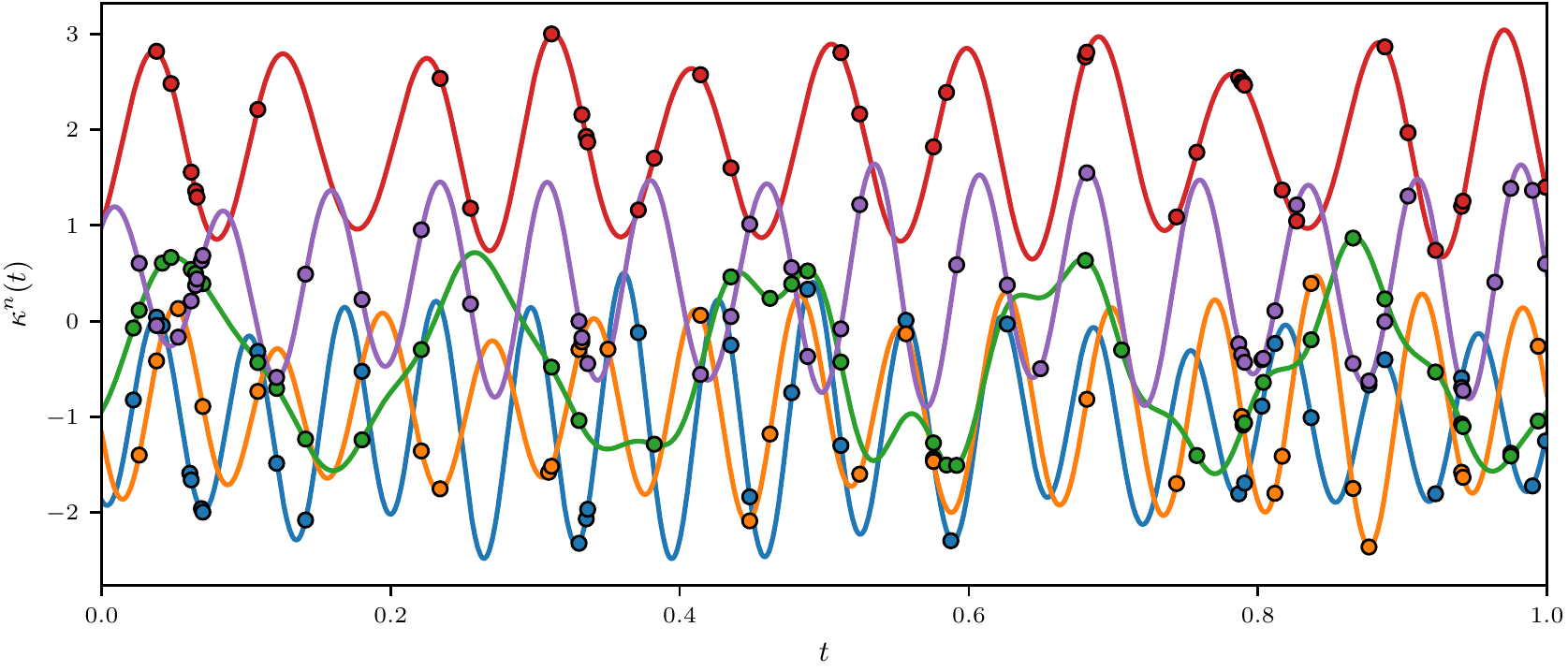}
\caption{
Examples of signals in the synthetic periodic data for 5 of the 20 nodes. 
The lines show clean signals $\simsignal{t}$ and each dot a (noisy) observation $y^n(t_i)$.
Note that the time steps are irregular and that not all signals are observed at each observation time.
}
\label{fig:periodic_signals}
\end{center}
\end{figure*}

Based on this graph 200 irregular time series are sampled according to Eq. \ref{eq:base_signal} and \ref{eq:synth_signal}.
An example is shown in \reffig{periodic_signals}.
The irregular time steps are created by first discretizing $[0,1]$ into 1000 time steps and then sampling 70 of these independently for each time series.
Out of all node observations we keep 50\%, sampled in the same way as for the traffic data.
The parameters of the periodic signals are sampled according to $\phi^n \sim \uniform{[20,100]}$ and $\eta^n \sim \uniform{[0, 2\pi]}$.
We resample all $\eta^n$ for each sequence, but sample the node frequencies $\phi^n$ only once, treating these as underlying properties of the nodes. 
Out of the 200 sampled sequences we use 100 for training, 50 for validation and 50 for testing.

\section{DETAILS ON EXPERIMENT SETUPS}
\label{sec:experiment_details}
We perform hyperparameter tuning by exhaustive grid search over combinations of parameter values.
The configuration that achieves the lowest validation $\lmse$ is then used for the final experiment.
An overview of hyperparameter values considered and the final configurations for all experiments is given in \reftab{hyperparameters}.
All hyperparamter tuning for \grubase and \itgnn is done on the versions of the models with exponential dynamics.
For these models we generally observe that larger architectures perform better, but the memory usage limits how much we can increase the model size.
While the exact model architecture of \itgnn does impact the results, the model does not seem particularly sensitive to learning rate or batch size.

\begin{table}[tbp]
\centering
\caption{
Values considered in hyperparameter tuning for the different datasets.
Bold numbers represent the best performing configuration, that was then used in the final experiment.
}
\label{tab:hyperparameters}
\begin{tabular}{@{}lcccccc@{}}
\toprule
 &
  Learning rate &
  $\dh$ &
  \begin{tabular}[c]{@{}c@{}}Fully connected\\ layers in $g$\end{tabular} &
  \begin{tabular}[c]{@{}c@{}}\gls{GNN} layers\\ in $g$\end{tabular} &
  \begin{tabular}[c]{@{}c@{}}\gls{GNN} layers\\ in GRU\end{tabular} &
  Batch size \\ \midrule
\textbf{Traffic Data}  &                        &                            &               &               &               &    \\ \midrule
\grujoint              & \textbf{0.001}, 0.0005 & 64, 128, \textbf{256}, 512 & 1, \textbf{2} & -             & -             & \textbf{16} \\
\grunode               & \textbf{0.001}, 0.0005 & 64, 128, \textbf{256}      & 1, \textbf{2} & -             & -             & \textbf{8}  \\
\itgnn                 & \textbf{0.001}, 0.0005 & 64, 128, \textbf{256}      & 1, \textbf{2} & 1, \textbf{2} & 1, \textbf{2} & \textbf{8}  \\
\transjoint            & \textbf{0.001} & \textbf{64}, 256, 512, 2048     & \multicolumn{3}{c}{\textbf{2}, 4 (\transbase layers)} & \textbf{8}  \\ 
\transnode            & \textbf{0.001} & \textbf{64}, 128, 256      & \multicolumn{3}{c}{\textbf{2}, 4 (\transbase layers)} & \textbf{8}  \\ \midrule
\textbf{USHCN Data}    &                        &                            &               &               &               &    \\ \midrule
\grujoint              & \textbf{0.001}         & 64, 128, 256, \textbf{512} & \textbf{2}    & -             & -             & \textbf{16} \\
\grunode               & \textbf{0.001}         & 32, 64, \textbf{128}       & \textbf{2}    & -             & -             & \textbf{4}  \\
\itgnn                 & \textbf{0.001}         & 32, 64, \textbf{128}       & \textbf{2}    & 1, \textbf{2} & 1, \textbf{2} & \textbf{4}  \\
\transjoint            & \textbf{0.001} & \textbf{64}, 256, 512, 2048     & \multicolumn{3}{c}{2, \textbf{4} (\transbase layers)} & \textbf{8}  \\ 
\transnode            & \textbf{0.001} & \textbf{64}, 128      & \multicolumn{3}{c}{\textbf{2}, 4 (\transbase layers)} & \textbf{8}  \\ \midrule
\textbf{Periodic Data} &                        &                            &               &               &               &    \\ \midrule
\grunode               & \textbf{0.001}         & 64, \textbf{128}, 256      & \textbf{2}    & -             & -             & \textbf{16} \\
\itgnn                 & \textbf{0.001}         & 64, \textbf{128}, 256      & \textbf{2}    & \textbf{2}    & \textbf{2}    & \textbf{16} \\ 
\transjoint            & \textbf{0.001} & 64, \textbf{256}, 512, 2048     & \multicolumn{3}{c}{\textbf{2}, 4 (\transbase layers)} & \textbf{8}  \\ 
\transnode            & \textbf{0.001} & 64, \textbf{128}, 256      & \multicolumn{3}{c}{2, \textbf{4} (\transbase layers)} & \textbf{8}  \\ \bottomrule
\end{tabular}
\end{table}

We evaluate $\lmse$ on the validation set after each training epoch, stopping the training early if the validation error does not improve for 20 epochs. 
Except for \lgode we train all models until this early stopping occurs, which typically takes less than 150 epochs.

\subsection{Traffic Data}
For the traffic data we perform hyperparameter tuning on the versions of the datasets with 25\% observations.
The same hyperparameter configuration, shown in \reftab{hyperparameters}, performed the best for both \bay and \la.

We used the slightly downscaled version of the \lgode model for the traffic data.
We tried also the default hyperparameters on the \la dataset, but this did not improve the results.
\lgode was trained for 50 epochs with a batch size of 8 and we observed no further improvements when trying to train the model for longer.

\subsection{USHCN Climate Data}
For the \gls{USHCN} data we perform the hyperparameter tuning on the $\tmin$ dataset.
Since the \gls{USHCN} graph contains many nodes we are somewhat more restricted in how large we can scale up the models.

\subsection{Synthetic Periodic Data}
On the periodic data we tried the same hyperparameters for \grujoint as for the other models, but no options gave any useful results.
Because of this we exclude the model from this experiment.
The number of iterations until the validation $\lmse$ stops decreasing is higher for the periodic data, with models training up to 500 epochs.

We used default hyperparameters for the \lgode model on the periodic data, but with a batch size of 16.
As this graph only contains 20 nodes we were here able to train 5 \lgode models with different random seeds.

\subsection{Loss Weighting}
For the loss weighting experiment we do not perform any new hyperparameter tuning, but use the best configuration from the traffic data experiment.
One \itgnn model was trained for each weighting function.
We here use a batch size of 4 and $\ninit = 25$, which allows us to study predictions to higher $\Delta_t$ in the future.

\bibliographyapp{references}
\bibliographystyleapp{apalike}

\end{document}